\documentclass[10pt,journal,compsoc]{IEEEtran}
\usepackage{graphicx}
\usepackage{comment}
\usepackage{amsmath,amssymb} 
\usepackage{booktabs}

\usepackage{hyperref}

\ifCLASSOPTIONcompsoc
  \usepackage[nocompress]{cite}
\else
  \usepackage{cite}
\fi
\ifCLASSINFOpdf
\else
\fi
\begin{document}
%
\title{DeepFN: Towards Generalizable Facial Action Unit Recognition with Deep Face Normalization}
%
%
%
%

\author{Javier~Hernandez, 
        Daniel~McDuff, 
        Ognjen~(Oggi)~Rudovic, 
        Alberto~Fung, 
        and~Mary~Czerwinski

\IEEEcompsocitemizethanks{

\IEEEcompsocthanksitem J. Hernandez, D. McDuff, M. Czerwinski are with Microsoft Research.\protect\\
E-mail: javierh@microsoft.com
\IEEEcompsocthanksitem A. Fung was with the University of Houston.
\IEEEcompsocthanksitem O. Rudovic was with the Massachusetts Institute of Technology. 

}
}

\IEEEtitleabstractindextext{%
\begin{abstract}

Facial action unit recognition has many applications from market research to psychotherapy and from image captioning to entertainment. Despite its recent progress, deployment of these models has been impeded due to their limited generalization to unseen people and demographics. This work conducts an in-depth analysis of performance across several dimensions: individuals (40~subjects), genders (male and female), skin types (darker and lighter), and databases (BP4D and DISFA). To help suppress the variance in data, we use the notion of self-supervised denoising autoencoders to design a method for deep face normalization (DeepFN) that transfers facial expressions of different people onto a common facial template which is then used to train and evaluate facial action recognition models. We show that person-independent models yield significantly lower performance (55\% average F1 and accuracy across 40 subjects) than person-dependent models~(60.3\%), leading to a generalization gap of $\Delta$5.3\%. However, normalizing the data with the newly introduced DeepFN significantly increased the performance of person-independent models~(59.6\%), effectively reducing the gap. Similarly, we observed generalization gaps when considering gender~($\Delta$2.4\%), skin~type~($\Delta$5.3\%), and dataset~($\Delta$9.4\%), which were significantly reduced with the use of DeepFN. These findings represent an important step towards the creation of more generalizable facial action unit recognition systems.

\end{abstract}

}


\maketitle

\IEEEdisplaynontitleabstractindextext

%
\IEEEpeerreviewmaketitle

\IEEEraisesectionheading{\section{Introduction}\label{sec:introduction}}

\IEEEPARstart{F}{acial} expression recognition technology offers the opportunity to comfortably capture the expressed emotional experience of people and facilitate unique interaction experiences~\cite{Martinez2019}. While the specific meaning of facial expressions may vary depending on the context~\cite{barrett2019emotional}, these signals have been successfully used in a wide variety of settings such as promoting engagement in human-robot interaction~\cite{Rudovic2018a,Rudovic2018b}, monitoring depression of patients~\cite{Cohn2009,stratou2013automatic}, estimating experienced pain~\cite{Lucey2011,kaltwang2012continuous,martinez2017automatic}, measuring engagement of TV viewers~\cite{McDuff2015,Hernandez2013b}, and promoting driver safety~\cite{Assari2011,Gao2014} among others.

To help quantify facial activity, researchers often rely on the Facial Action unit Coding System (a.k.a.,~FACS)~\cite{ekman1997face,Cohn2007} which decomposes facial movements into different muscle activations called action units (AUs). For instance, AU12 indicates the activation of the zygomaticus major muscle which pulls the corner of the lip and is frequently seen during smiles. However, obtaining high quality labels for each person can be very expensive and time consuming. For instance, FACS labels are usually provided by expert coders which may spend up to 30 minutes to label around one-minute of video data~\cite{zhao2018learning}. To help provide scalable FACS labeling, researchers have proposed and developed a wide variety of methods and tools that automatically detect AUs from face images~(e.g.,~\cite{li2019semantic,  baltrusaitis2018openface, mcduff2019multimodal, Corneanu2016, Martinez2019, Li2020}). 

Despite the significant advancements in the recent years, deploying such models in the wild is still challenging. One of the biggest hurdles, which also applies to most human-centered AI applications, involves the development of models that can generalize well across individuals or groups of people despite their differences. In the context of facial action unit recognition, some of the most readily observable differences are those associated with individual facial appearance such as  head shape, amount of facial hair or skin type~which can significantly vary across people and impair the ability of models to recognize relevant facial activity. As a consequence, machine learning models trained/tested with (non-overlapping) data of the same person (a.k.a.,~person-dependent models) usually outperform those trained/tested on data from different people (a.k.a.,~person-independent models)~\cite{Cohen2003,Valstar2011}. Similar differences have been observed when considering group-independent and dependent models that split groups according to different criterion that may similarly impact facial appearances~(e.g.,~gender~\cite{Fink2005}, skin type~\cite{Zhuang2010}). 

Poor cross-group generalization performance of machine learning models can partly be explained due to the violation of the independent and identically  distributed (iid) data assumption, since the training/test data are sampled from highly correlated data (multiple images of the same person) and data with non-stationary distribution (multiple persons and over different periods of time). This is particularly pronounced when considering certain groups of people who may be underrepresented in existing training sets (e.g.,~older age, darker skin)~\cite{buolamwini2017gender,buolamwini2018gender,wilson2019predictive}. Interestingly, humans have been shown to be similarly impaired with the out-group homogeneity bias~\cite{Quattrone1980} and the cross-race effect~\cite{Elfenbein2002} -- suggesting that people are usually better at identifying facial variation of in-group members versus out-across members as well. Furthermore, people are usually better at recognizing emotions associated with expressions in faces from people within their own demographics group~\cite{Elfenbein2002}. In both cases, the underlying biases in the data itself may result in poor decision making and the reinforcement of harmful stereotypes~\cite{buolamwini2017gender}.  


Motivated by the previous limitations, 
this work explores whether we can minimize appearance differences across people while maintaining their facial expressions with the goal of improving cross-group generalization in the context of AU recognition. The contributions of this work are as follows.
\begin{enumerate}
	\item This is the first work to systematically examine and compare cross-group generalization across multiple group splits within the same experimental conditions. To do so, we evaluated group-dependent and independent models based on individuals (40~people), genders (male and female), skin types (light and dark), and datasets (BP4D and DISFA) which capture some of the typical sources of variance of real-life deployments. Among the different conditions, we found cross-dataset generalization to be the most challenging one.
	\item This is the first work to explore the use of face transfer in the context of face normalization. In particular, we propose DeepFN which leverages self-supervised denoising autoencoders to transfer facial expressions of different people onto a common facial template (a.k.a.,~template of reference). 
	This simple yet effective method was successful in significantly closing the gap between group-dependent and independent models, especially when considering different skin types and genders. 
	\item In light of our findings, we discuss the limitations of this work and draw  important ethical considerations around reducing algorithmic biases, facilitating AI interpretability, and preserving data privacy to help guide future research efforts in this area. 
\end{enumerate}



The remainder of the paper is organized as follows. First, we review prior work on data normalization methods in the context of facial expression recognition. Second, we describe the proposed methodology which includes the  appearance normalization, the template selection, and the facial action unit recognition model. Third, we describe the experimental protocol which includes the considered groups, datasets, and evaluation. Fourth, we provide the results for each one of the considered groups. Finally, we discuss the results, limitations of the work, and ethical considerations. 



\begin{figure*}[t]
  \centering
  \includegraphics[width=1\linewidth]{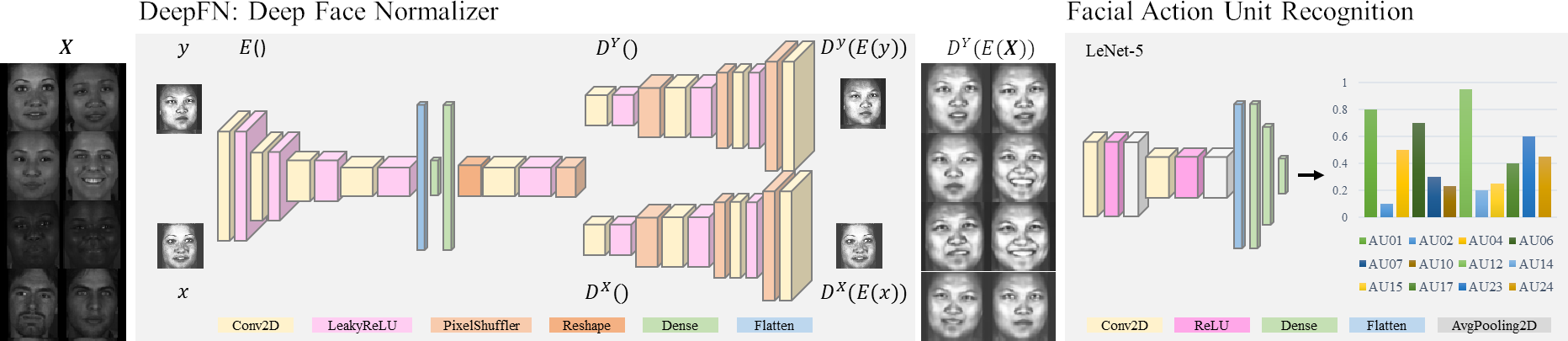}
  \caption{Overview of main the processing phases: 1)~face images are first normalized with DeepFN to ensure individual differences are minimized, and 2)~the normalized images are used as input into a LeNet-5 convolutional neural network to recognize 12 facial action units. }
  \label{fig:hero}
\end{figure*}

\section{Related Work}
Performance differences between person-dependent and person-independent models have been reported in several studies~\cite{Cohen2003,Valstar2011,Chen2013}. One the one hand, person-dependent models tend to perform better but offer limited practical use as obtaining high quality annotations for each person can be very expensive and time consuming. 
On the other hand, person-independent models tend to perform worse but they are more scalable as they do not require the training data from target (test) subjects. To help improve performance while still maintaining scalability, researchers have explored a wide variety of methods to help minimize individual differences in the context of person-independent models.

Arguably one of the most simple and commonly explored normalization methods involves applying some sort of intuitive correction on the face images or sensed features that helps minimize differences across people. For instance, several studies have considered using the distance between the eyes to help correct geometric or shape-based features such as distances between different facial landmarks~\cite{Tian2001,hernandez2013measuring,Niu2019}. 
Similarly, other studies have considered computing differences across different regions of the face to help correct appearance-based features such as texture~\cite{Niu2019} or applying color correction methods such as histogram equalization to reduce the data distribution mismatch in the color space~\cite{Tran2017}. 
Another correction method involves the use of face alignment, especially in the context of face recognition. For instance, Banerjee~et~al.~\cite{banerjee2018frontalize} validated recent 2D/3D face-alignment methods to quantify the effect of this pre-processing step in the context of face recognition. As noted by the authors, however, the quality of alignment did not directly translate into the performance of the face-recognizers, as various (individual) facial details could be lost during this processing step. This may be even more critical in the case of facial action recognition as facial appearance and dynamics of AUs are very person-specific. The work presented in this paper similarly explores correcting the input images but explores using more complex methods to normalize facial appearance. In addition, we found that pre-processing the images with histogram equalization helped facilitate the normalization process. 

An alternative normalization method involves the use of unlabeled testing data to help calibrate the methods. 
For instance, Baltrusaitis~et~al.~\cite{Baltrusaitis2015} explored using unlabeled testing data to estimate the neutral-looking face image and serve as a reference to capture potential activations. While this led to successful calibration of the AU classifiers, this method depends on tracking methods such as the Constrained Local Models (CLMs) which are highly sensitive to the performance of the facial landmark detection. 
Other works have explored the smart selection or weighting of training samples to ensure that they are representative of the testing samples. For instance, Chu~et~al, ~\cite{Chu2017} proposed a Selective Transfer Machine (STM) approach for personalized facial expression analysis - an inductive learning approach that tries to align the distributions of the facial features' of training and test subjects during training. 
In a separate effort, Zen~et~al.~\cite{zen2014unsupervised} proposed to learn a regression function that would learn from training subjects the optimal parameters of a facial expression classifier for a specific person. 
Similarly, Feffer~et~al.~\cite{feffer2018mixture} proposed a Mixture-of-Experts deep learning approach that would tackle the individual differences by selecting the most similar training subjects and their corresponding expert-classifiers for the task of valence/arousal estimation. 
Different from these works, Yang~et~al.~\cite{yang2014personalized} proposed a two-step approach for personalized modeling of facial AU intensity from spontaneously displayed facial expressions. In the first step, a matrix decomposition algorithm (unsupervised) was applied to separate the identity from facial expression of target subjects. The obtained representations (expression plus identity features) were then jointly modeled using the framework of Conditional Ordinal Random Fields, resulting in a personalized model for intensity estimation of AUs. The work presented in this paper similarly requires the use of unlabeled testing data but differs from previous efforts in several important ways: 1) the normalization step is completely separated from the task of facial action unit recognition task allowing easier optimization, 2) the normalization is performed at the image level allowing easier inspection of the output data, and 3) the normalization does not require the use of facial landmark detection reducing the dependency on other methods. Due to these, the proposed method could be  used as a pre-processing step for any of the facial action unit recognition methods. To start exploring the potential value of the proposed method, however, we selected a simple and lightweight LeNet-5 which allowed quickly evaluation of generalization performance across different groups. 

While the previous methods mainly focus on variations in facial appearance and features, it is worth noting that researchers have also investigated other methods to normalize other sources of variation such as out-of-plane head rotations, illumination changes, and other (typically) external artifacts that may be in the data~\cite{qian2019unsupervised, haghighat2016fully, sagonas2017robust}. In the case of large-out-of-plane head-rotations, the majority of studies attempt face frontalization - effectively mapping the non-frontal faces to a frontal reference frame. For instance, Werner~et~al.~\cite{Werner2019} proposed a face normalization method which also depends on the quality of the facial landmark detection and texture-warping. The authors showed that with their face-normalization method they could train better-performing CNN models for facial expression recognition and AU detection, compared to when no face-normalization was applied. 
The work presented here focuses on minimizing differences in facial appearance across target people while preserving information about their facial expressions. Even though there are examples of other sources of variation (such as the extreme head poses) in data used in this work, the majority of data falls under the in-plane head rotations. We show empirically that the proposed method based on self-supervised autoencoders can successfully deal with this range of head pose variation without adversely affecting the AU estimation performance.



\section{Methods}

This section describes the methodological details of the proposed approach, DeepFN, as well as the facial action unit recognition algorithm.

\subsection{Facial Appearance Normalization}
Some of the most popular methods for facial expression transfer and expression synthesis start by detecting facial landmarks or action units to help guide the transfer process (e.g.,~\cite{Thies2015,Kim2018,Zakharov2019,Nirkin2019,Mirsky2020}). As this work explores ways to improve the task of facial action unit recognition, we chose a method that does not require any explicit indication of facial landmarks or action units. In particular, we selected a self-supervised autoencoder approach which is composed of three main components: (i)~a shared encoder network ($E$), used to reduce the dimensionality of face images onto a lower dimensional latent space, (ii)~one decoder network ($D^y$), used to recover images of the reference template selected for the normalization, and (iii)~a second decoder network, used to recover images of the individual, the face image of whom we wish to normalize ($D^x$). Figure~\ref{fig:hero}~(left) illustrates the main components and how they are combined.

During training, the face images~(the reference template and the image of the subject to be normalized) were iteratively compressed with the encoder and recovered with their corresponding decoder, while minimizing the root mean squared error~(RMSE) between the original input and recovered image pixel values. As the same encoder is used to embed the images of the two appearances, the latent space learns to capture the sources of variance shared by the two inputs~(e.g.,~head poses and facial expressions). 
As a preprocessing step, all the input images were converted to gray scale and the pixels values were normalized using the histogram equalization approach in order to facilitate a more consistent distribution of pixel values across individuals, and also remove artifacts such as illumination changes, skin colour variation, etc. In addition, random affine transformations and Gaussian warps were used as image augmentation techniques to help increase the amount of training data~(and thus, the variation of input images). Specifically, the training process involved optimizing the following joint self-supervised loss function:


\begin{equation} \label{eq:1}	
  \arg\min_{E,D^x,D^y} \frac{1}{n} \sum_{i=1}^n |x_i - D^x (E(x_{i}^{'}))| + \frac{1}{m} \sum_{j=1}^n |y_j - D^y (E(y_{j}^{'}))|
 \end{equation}

where $x_i$ represents the $i$-th image of the person to be normalized, $y_j$ represents the $j$-th image of the template of reference, and $x_{i}^{'}$ and $y_{j}^{'}$ represent their pre-processed versions, respectively. This process was repeated for each person that needed to be normalized.


During the testing phase, the images of the individual to be normalized were similarly compressed (without augmentation) but recovered with the alternate decoder ($D^y$) as follows:
\begin{equation} \label{eq:3}	
    X^{y}  = D^y (E(X))
\end{equation}
where $X^y$ represents the normalized images of the person.



Figure~\ref{fig:hero} shows the deep neural network architecture used which was built on top of an existing code base\footnote{\url{https://github.com/joshua-wu/deepfakes\_faceswap}}. To ensure appropriate transfer, we allowed the network to learn for 50K iterations which included a batch of 64 randomly selected template images followed by another batch of 64 randomly selected images from the person to be normalized. Table~\ref{table:encoder} and Table~\ref{table:decoder} show the specific architecture implementation for the encoder and decoders, respectively.

\begin{table}[t]
    \begin{center}
            \caption{Encoder network architecture used to compress facial information.}
        \begin{tabular}{ccc}
            \toprule
            \textbf{Layer} & \textbf{Filters, Kernel Size, Strides} & \textbf{Output}\\ \hline \hline
            Input & - & 128, 128, 1\\ 
            Conv2D/LeakyReLU & 128, 5, 2 & 64, 64, 128\\ 
            Conv2D/LeakyReLU & 256, 5, 2 & 32, 32, 256\\ 
            Conv2D/LeakyReLU & 512, 5, 2 & 16, 16, 512\\ 
            Conv2D/LeakyReLU & 1024, 5, 2 & 8, 8, 1024\\ 
            Flatten &- & 65536\\ 
            Dense & - & 1024\\ 
            Dense & - & 16384\\ 
            Reshape & - & 4, 4, 1024\\ 
            Conv2D/LeakyReLU & 2048, 3, - & 4, 4, 2048\\ 
            PixelShuffler & - & 8, 8, 512\\ 
            \bottomrule
            \label{table:encoder}
        \end{tabular}
    \end{center}
\end{table}

\begin{table}[t]
    \begin{center}
            \caption{Decoder network architecture used to recover facial information.}
        \begin{tabular}{ccc}
            \toprule
            \textbf{Layer} & \textbf{Filters, Kernel Size, Strides} & \textbf{Output}\\ \hline \hline
            Input & - & 8, 8, 512\\ 
            Conv2D/LeakyReLU &1024, 3, - & 8, 8, 1024\\ 
            PixelShuffler & - &  16, 16, 256\\             
            Conv2D/LeakyReLU & 512, 3, - & 16, 16, 512\\ 
            PixelShuffler & - & 32, 32, 128\\ 
            Conv2D/LeakyReLU & 256, 3, - & 32, 32, 256\\ 
            PixelShuffler & - & 64, 64, 64\\ 
            Conv2D/LeakyReLU & 128, 3, - & 64, 64, 128\\ 
            PixelShuffler & - & 128, 128, 32\\   
            Conv2D/Sigmoid & 1,5,- & 128, 128, 1\\ 
            \bottomrule
            \label{table:decoder}
        \end{tabular}
    \end{center}
\end{table}

\begin{table}
    \begin{center}
            \caption{LeNet-5 architecture used for facial action unit recognition.}
        \begin{tabular}{ccc}
            \toprule
            \textbf{Layer} & \textbf{Filters, Kernel Size} & \textbf{Output}\\ \hline \hline
            Input & - & 64, 64, 1\\ 
            Conv2D/ReLu & 6, 3x3 & 62, 62, 6\\ 
            AvgPool2D & - & 31, 31, 6\\ 
            Conv2D/ReLu & 16, 3x3 & 29, 29, 16\\ 
            AvgPooling2D & - & 14, 14, 16\\ 
            Flatten &- & 3136\\ 
            Dense & - & 120\\ 
            Dense & - & 84\\ 
            Dense/Sigmoid & -  & 12 or 5\\ 
            \bottomrule
            \label{table:lenet}
        \end{tabular}
    \end{center}
\end{table}

\subsection{Template Selection}
To normalize the facial appearance across people, DeepFN requires a template ($Y$) that provides the appearance to be shared.  While there are many potential options for the selection of the template, this work leveraged the images of the most expressive subject of the BP4D dataset which was more likely to capture a rich gamut of facial variations while still resembling some of the main dataset characteristics (e.g.,~camera angle, illumination). In particular, we  first counted the number of facial action unit activations for each person, and then selected the participant for which the median across all action units was the highest in the dataset (see~$y$ in Figure~\ref{fig:hero} and bottom row of Figure~\ref{fig:avg_face}). This participant was excluded from the action unit classification analysis to help provide a fair comparison. While not included in this work, we also explored the creation of synthetic reference models such as avatars that can be easily controlled~\cite{VanderStruijk2018,Aneja2019} or synthetic methods such as StyleGAN~\cite{Karras2019}. However, we found that the demographics as well as range of facial expressions were limited compared to when considering real subjects~\cite{Salminen2020}, lowering the quality of the expressions transfer results. 

\subsection{Action Unit Recognition}

Once all the images were transferred into a common template of reference, we fed them into a facial action unit classifier that only sees a single appearance (the images mapped to the template). For the purpose of our analysis, we selected the LeNet-5 convolutional neural network architecture (see~right of Figure~\ref{fig:hero}) which was originally proposed by LeCun~et~al.~\cite{LeCun1998}. Among the different possibilities, we selected LeNet-5 as it provided a simple and lightweight solution (388K parameters) to quickly evaluate different experimental conditions while avoiding overfitting. 
In terms of loss function, we used the binary cross entropy function which allows for multiple probabilistic activations for a single image. Table~\ref{table:lenet} shows the specific architecture implementation of the classifier. 

\section{Experimental Protocol}

\subsection{Groups}
The main motivation of this work is to improve the generalization of facial action unit models across different groups of people. To help evaluate this, we performed multiple within-group and cross-group evaluations across different group splits, 
and then applied DeepFN to evaluate whether it helped bridge the performance gap. 
In particular, we considered the following group splits.


\textbf{Person}. The first group split considers each individual as their own separate group as it captures the most frequently considered source of facial appearance variance (e.g.,~\cite{Cohen2003,Valstar2011}), especially in the context of face recognition~\cite{Qian2019}. The within-group evaluations included models that were trained and tested with data from the same person (a.k.a.,~person-dependent models). The cross-group evaluations included models that were trained and tested with data from different people (a.k.a.,~person-independent models). In this case, person-dependent models capture the optimal performing scenario in which identity labels and data from a specific subject are available and, consequently, good model generalization can be more easily achieved.

\textbf{Gender}. The second group split divides participants according to their gender (male and female) which captures facial variance in terms of sex-related facial characteristics such as the amount of hair or the shape of the jaw~\cite{Fink2005}. 
The within-group evaluations included models that were trained and tested with male participants as well as models that were trained and tested with female participants. The cross-group evaluations included models that were trained  with males and tested with females, as well as the opposite. For convenience, we will refer to these experiments as gender-dependent and gender-independent models, respectively. However, it is important to note that both of them fall under the category of person-independent models as the subjects used for training and validation were different than the ones used for testing.

\textbf{Skin Type}. The third group split divides participants according to their skin skin type~(lighter and darker skins) which captures facial variance in terms of skin pigmentation and other facial characteristics that may be correlated correlated with it (e.g.,~shape of the nose~\cite{Zhuang2010}). This group split has been more frequently been examined in the context of algorithmic biases of computer vision~\cite{buolamwini2017gender,buolamwini2018gender, wilson2019predictive}. The within-group evaluations included models that were trained and tested with participants with lighter skin (e.g.,~skin type~I and II) as well as models that were trained and tested with participants with darker skin (e.g.,~skin type~V and~VI) (a.k.a.,~skin-dependent models). The cross-group evaluations included models that were trained with participants with lighter skin and tested with those with darker skin, as well as the opposite (a.k.a.,~skin-independent models). To annotate skin type~for each of the participants, we used the Fitzpatrick Phototype~Scale~\cite{fitzpatrick1988validity} which separates skin types into six main categories. To help amplify differences associated with skin types, we group participants into lighter skin (types below or equal to~II) and darker skin (types above or equal to V). 

\textbf{Dataset}. The fourth group split divides participants according to the dataset in which they participated (BP4D and DISFA) which captures facial variance in terms of a wide variety of factors such participant demographics and specific data collection settings (e.g.,~camera angle, illumination)~\cite{Baltrusaitis2015}. The within-group evaluations included models that were trained and tested with participants from the same dataset (a.k.a.,~database-dependent models). The cross-group evaluations included models that were trained  with one dataset and tested with those from another one (a.k.a.,~database-independent models).

\subsection{Datasets}

\begin{figure}[t]
  \centering
  \includegraphics[width=1\linewidth]{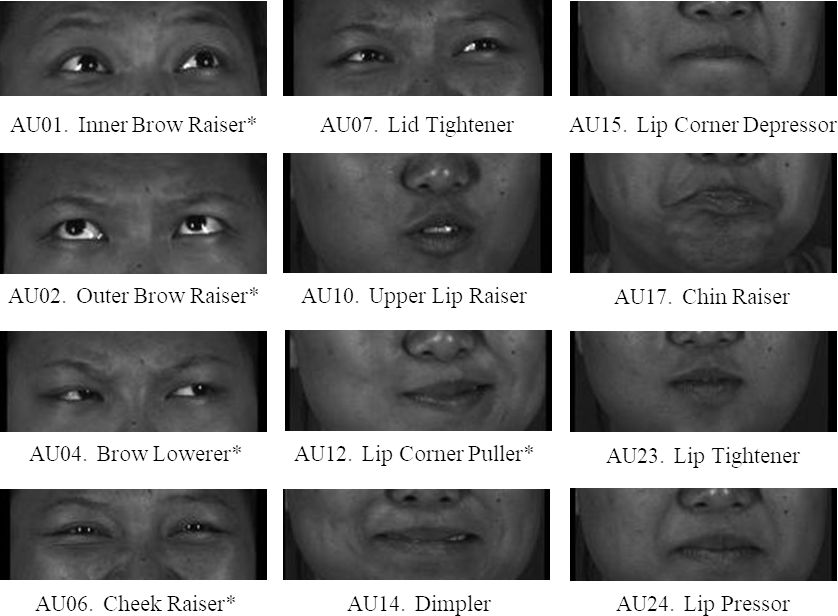}
  \caption{Subset of facial action units~\cite{Kanade2000, Cohn2007} considered in the study. *~indicates the actions used when studying cross-dataset generalization.}
  \label{fig:summary_aus}
\end{figure}

To systematically evaluate performance under the different group splits, this work mainly leverages the benchmark Binghamton-Pittsburgh 4D Spontaneous Expression Dataset (BP4DSpontaneous)~\cite{Zhang2014} due to its diverse set of demographics and wide use in the context of facial action unit recognition. In particular, the dataset contains a total of around 140K images distributed across 41 participants (23~females) who were instructed to perform 8 different tasks designed to elicit authentic emotions (e.g.,~playing games, watching a film, social interview). The images were annotated by expert FACS coders. In our study, we focused on the following 12 facial action units: AU01 (inner brow raiser), AU02 (outer brow raiser), AU04 (brow lowerer), AU06 (cheek raiser), AU07 (lid tightener), AU10 (upper lip raiser), AU12 (lip corner puller), AU14 (dimpler), AU15 (lip corner depressor), AU17 (chin raiser), AU23 (lip tightener), and AU24 (lip pressor). In addition to providing binary occurrence values for each of the actions units, the dataset also includes intensity values (ranging from 1 to 8) for a smaller subset of the action units (AU06, AU10, AU12, AU14 and AU17) which will be considered in part of the analysis. Figure~\ref{fig:summary_aus} shows visual examples of the different action units.

To study cross-dataset generalization, we also use the Denver Intensity of Spontaneous Facial Actions (DISFA) benchmark dataset~\cite{Mavadati2013} which includes around 130K images distributed across 27 participants (12 females) watching emotive video stimulus. Similarly, the images were annotated by expert coders. For the purpose of our experiments, we focused on the following five facial action units which were also provided in the BP4D dataset: AU01, AU02, AU04, AU06, and AU12. See~*~in~Figure~\ref{fig:summary_aus}.

\begin{figure*}[t]
  \centering
  \includegraphics[width=1\linewidth]{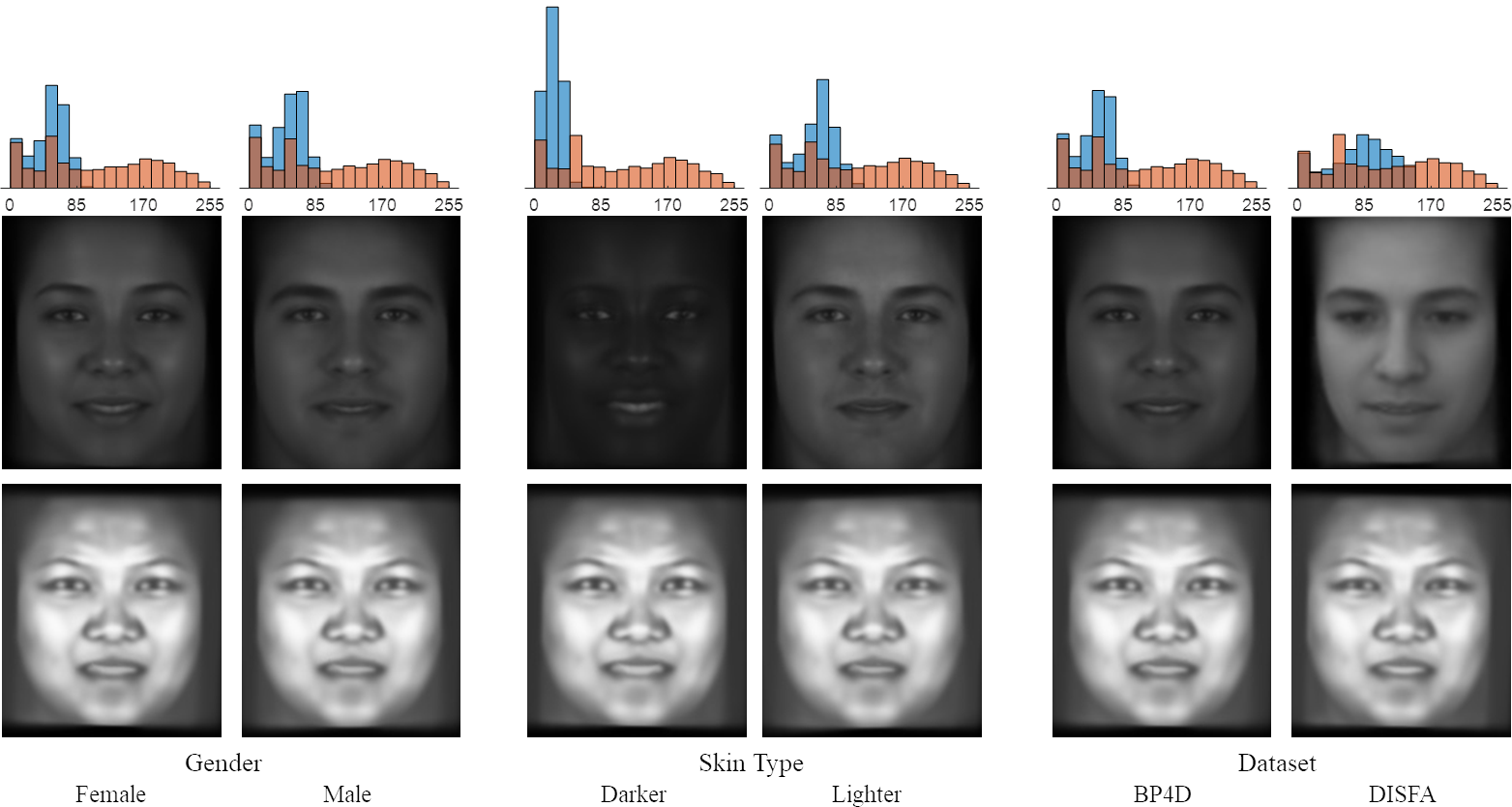}
  \caption{Average facial appearance before (middle row) and after applying DeepFN normalization (bottom row) for different groups. The top row shows the luminance histograms of the average images before (blue) and after (orange) normalization.}
  \label{fig:avg_face}
\end{figure*}

Table~\ref{tab:size_splits} shows the number of participants considered for each of the group splits discussed in the previous section. In particular, there were seven participants that met the criteria of having lighter skin type~(two females), eight with darker skin type~(six females), 22~female, 18~males, 40~in BP4D, and 27~in DISFA. To help illustrate appearance differences across different group splits, Figure~\ref{fig:avg_face} shows the average facial appearance (middle row) as well as their luminance histograms (blue bars on the top row). The figure also includes the same information after applying DeepFN (bottom row and red bars on the histogram) showing greater consistency across the different group splits.

\subsection{Evaluation}
Table~\ref{tab:overview_results} shows a summary of all the experiments. For each of the conditions, we randomly selected four participants for training, two for validation, and the remaining participants for testing from their corresponding pool of participants (see~Table~\ref{tab:size_splits}). These numbers were mostly determined by the size of the smaller group (seven people with lighter skin) and were kept the same across conditions to help facilitate a fair performance comparison. 
Each model was trained for a total of 50 epochs and testing predictions were obtained using the model that yielded the highest F1-score in the validation set. 
Each of the conditions was repeated 20 times to help minimize potential selection effects. All the neural networks were optimized with Adam with a learning rate of~0.00005 and exponential decay rates of 0.5 and 0.999 for the first and second-moment estimates, respectively. The specific implementation was built in Python~3.5.6, Keras~2.2.4, and Tensorflow~1.14.0. 

To capture the overall performance for each model, we first computed the average between the F1-score and accuracy for each of the action units (at a threshold of 0.5), and then aggregated them for each of the participants. For each of the conditions, we then computed the average and standard deviation across all the participants. To compare performance across conditions, we used two-sample t-test comparisons with a significance score when $p<0.05$.



\begin{table}[t]
  
    \caption{Number of available participants for each for the data splits}
    \label{tab:size_splits}
    \centering

  \begin{tabular}{p{16mm}c}
    \toprule
    \textbf{Group Split}&
    \textbf{Sample Size} \\ \hline \hline    
    People  &  40  \\  
    Female  &  22  \\  
    Male  &  18  \\  
    Darker skin  &  8  \\  
    Lighter skin  &  7  \\  
    BP4D  &  40  \\  
    DISFA  &  27  \\ \bottomrule  
    \end{tabular}%

\end{table}

\begin{table*}[t]
  
    \caption{(Top) Average and standard error bars for different experimental conditions and (bottom) overview of experimental details for each of the conditions including average accuracy and F1 score (\%) and standard deviation. \#AUs: number of facial action units, Reps.: repetitions.}
    \label{tab:overview_results}
    \centering
  \begin{tabular}{p{30mm}p{14mm}p{14mm}clp{25mm}ccc}
  \multicolumn{9}{l}{\includegraphics[width=.90\textwidth]{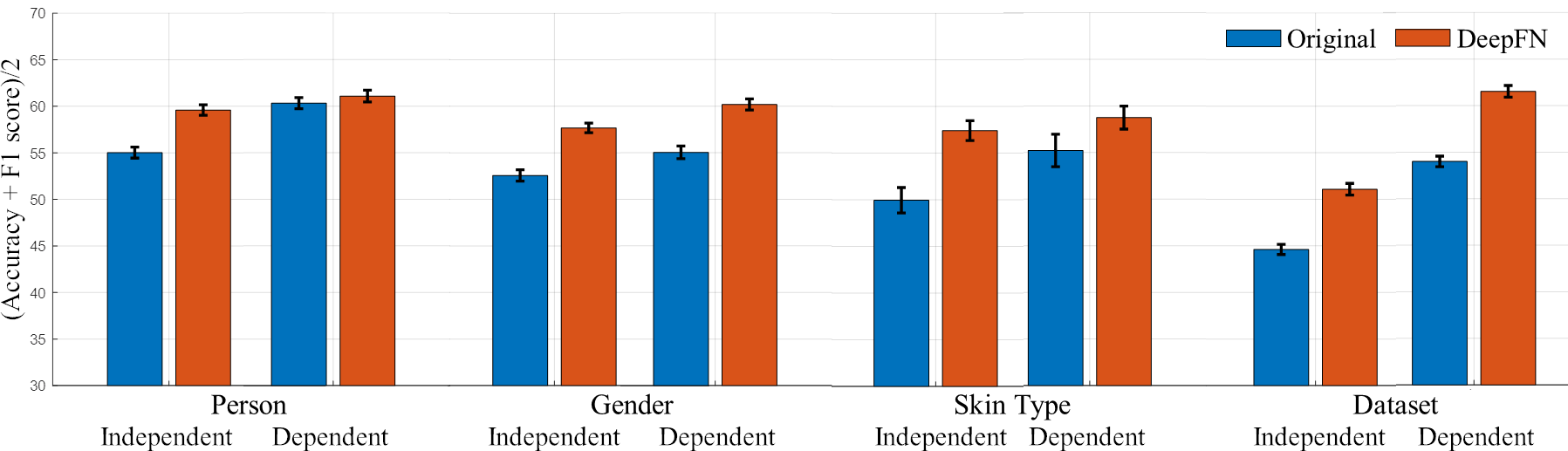}} \\
    \hline \hline 
    \textbf{Condition Name}&
    \textbf{Group} & 
    \textbf{Dataset} & 
    \textbf{\#AUs} & 
    \textbf{Analysis} & 
    \textbf{Train $\rightarrow$ Test} & 
    \textbf{Reps.} & 
    \textbf{Original} & 
    \textbf{DeepFN}  \\ \hline \hline    
    Person-dependent\newline & Person & BP4D & 12 & Within & Tasks $\rightarrow$ Tasks & 20x40 & 60.3 (3.8) & 61.1 (4.0)  \\\hline  
    Person-independent\newline & Person & BP4D & 12 & Cross & People $\rightarrow$ People & 20x2 & 55.0 (3.7) & 59.6 (3.6)  \\\hline  
    
    Gender-dependent & Gender & BP4D & 12 & Within &  Male $\rightarrow$ Male \newline Female $\rightarrow$ Female & 20x2 & 55.0 (4.3) & 60.2 (3.8)  \\\hline  
    Gender-independent & Gender & BP4D & 12 & Cross &  Male $\rightarrow$ Female \newline Female $\rightarrow$ Male & 20x2 & 52.6 (3.9) & 57.7 (3.2) \\\hline  
    
    Skin-dependent & Skin type~& BP4D & 12 & Within & Lighter $\rightarrow$ Lighter \newline Darker $\rightarrow$ Darker & 20x2 & 55.2 (6.5) & 58.7 (4.6)  \\\hline  
    Skin-independent & Skin type~& BP4D & 12 & Cross &  Lighter $\rightarrow$ Darker \newline Darker $\rightarrow$ Lighter & 20x2 & 49.9 (5.1) & 57.4 (4.0)  \\\hline 
    
    Dataset-dependent & Dataset & BP4D \newline DISFA & 5 & Within &  BP4D $\rightarrow$ BP4D \newline DISFA $\rightarrow$ DISFA & 20x2 & 54.0 (4.7) & 61.5 (5.1)  \\\hline  
    Dataset-independent & Dataset & BP4D \newline DISFA & 5 & Cross &  BP4D $\rightarrow$ DISFA \newline DISFA $\rightarrow$ BP4D & 20x2 & 44.6 (4.8) & 51.1 (5.1)  \\\hline  
    \end{tabular}%
    
\end{table*}



\begin{figure*}[t]
  \centering
  \includegraphics[width=1\linewidth]{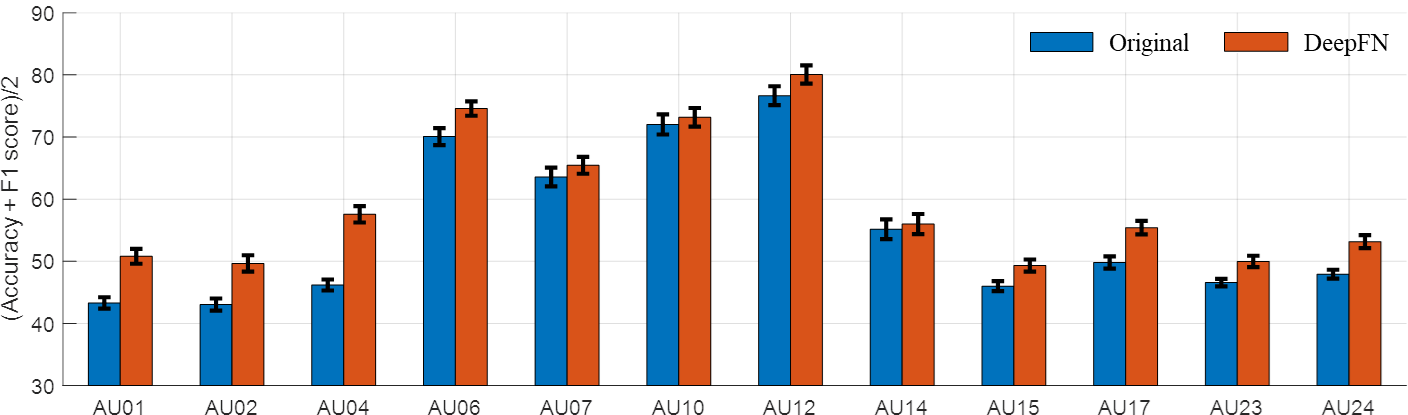}
  \caption{Average and standard error for person-independent models with (DeepFN) and without (original) normalization across facial action units.}
  \label{fig:au_results}
\end{figure*}

\section{Results}
Table~\ref{tab:overview_results} provides an overview of the results obtained for each of the experiments across conditions. In this section we will review the results for each of these groups.


\subsection{Cross-person Generalization}
\label{subsec:person}

When evaluating the models with the original (unnormalized) data, person-independent models achieved an average score of 55\% and person-dependent models achieved an average score of 60.3\% which were significantly different (\mbox{$p < 0.001$}). This difference indicates that the performance impact of having access to person-specific data is around 5.3\% in this dataset. This finding is consistent with previous work showing that person-dependent models lead to greater performance as both training and testing data follow the same distribution. In this case, 60.3\% can be considered as the optimal performing results for our experimental setting in which no individual differences exists.

When evaluating the models with DeepFN (normalized) data, we observe that person-independent models increased to 59.6\% which was significantly higher than its unnormalized person-independent counterpart (\mbox{$p < 0.001$}) and very similar to the unnormalized person-dependent results (\mbox{$p = 0.375$}). This finding suggests that DeepFN can effectively minimize individual differences associated with appearance. Person-dependent models with DeepFN maintained a performance of 61.4\% which was similar than its unnormalized counterpart (\mbox{$p = 0.388$}), suggesting that the face transfer process did not lose relevant facial expression information. 

Figure~\ref{fig:au_results} shows the average performance per action unit when using person-independent models with and without DeepFN. As can be seen, DeepFN provided an average improvement of around 4.6\% (STD: 2.9) for the considered AU with the greatest gains for AU04 (11.4\%) and the smallest gains for AU14 (0.8\%).

\subsection{Cross-gender Generalization}

When using the original data, gender-independent models achieved an average score of 52.6\% and gender-dependent models achieved an average score of 55\% which were significantly different (\mbox{$p = 0.009$}). This difference indicates that the impact of having different genders across training and testing sets is around 2.4\% in this dataset.

When using DeepFN, we observe that gender-independent models increased to 57.7\% which was significantly higher than its unnormalized counterpart (\mbox{$p < 0.001$}) and gender-dependent models increased to 60.2\% which was also significantly higher than its unnormalized counterpart (\mbox{$p < 0.001$}). The fact that gender-independent models with DeepFN yielded higher results than gender-dependent models without DeepFN indicates that the normalization is helping address individual differences beyond gender which is to be expected as the proposed method normalizes differences at the individual level.

\begin{figure}[t]
  \centering
  \includegraphics[width=1\linewidth]{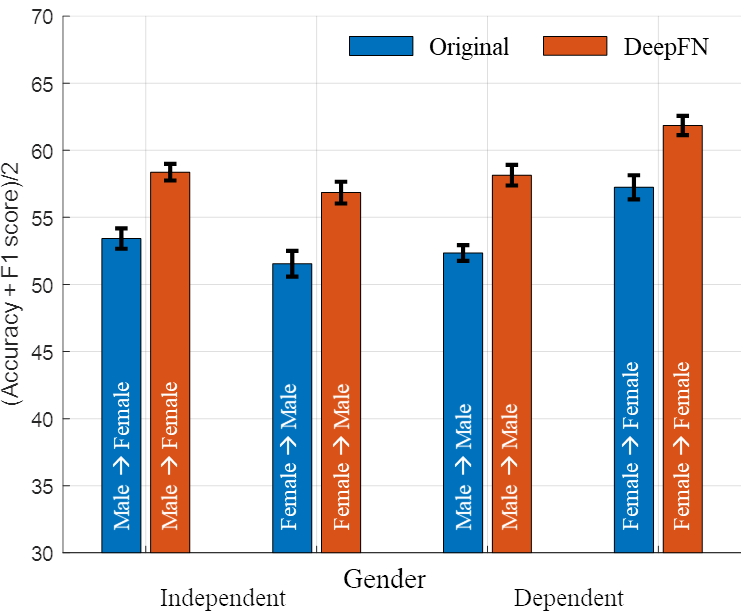}
  \caption{Average and standard error bars for different training/testing splits of gender-independent and dependent models. }
  \label{fig:gender_splits}
\end{figure}

To further explore these results, Figure~\ref{fig:gender_splits} shows the results for each of the training/testing combinations with and without DeepFN. As can be seen, the use of DeepFN yielded a consistent average improvement of 5.2\% across the different conditions. While we originally hypothesized that within-group evaluations would outperform cross-group evaluations, however, we  observe that experiments in which women were used as part of the testing set yielded better performance than those in which men were used as part of the testing set, irrespective of the training group and normalization method (\mbox{$p <= 0.001$}). When considering the intensity of facial action units, we observe that the activations of female participants were more intense than those associated with male participants on average (\mbox{$p <= 0.001$}), suggesting that recognizing expressions in females may be easier.

\subsection{Cross-skin Generalization}


When using the original data, skin-independent models achieved an average score of 49.9\% and skin-dependent models achieved an average score of 55.2\% which were significantly different (\mbox{$p = 0.025$}). This difference indicates that the impact of having different skin types across training and testing sets is around 5.3\% which is a bit larger than the generalization gap associated with gender (2.4\%). While we have not seen prior work comparing these two types of the generalization, this finding seems to suggest that skin type~may have a greater impact than gender in the context of model generalization. However, it is important to note that the number of subjects in the skin type~condition is smaller than in the gender condition (see~Table~\ref{tab:size_splits}). 

When using DeepFN, we observe that skin-independent models increased to 57.4\% which was significantly higher than its unnormalized counterpart (\mbox{$p < 0.001$}) and skin-dependent models increased to 58.7\% which was similar to its unnormalized counterpart (\mbox{$p = 0.110$}). In this case, skin-independent models with DeepFN yielded higher but comparable results than skin-dependent without DeepFN (\mbox{$p = 0.290$}), suggesting that the normalization method addressed the main source of data variance in this condition.

\begin{figure}[t]
  \centering
  \includegraphics[width=1\linewidth]{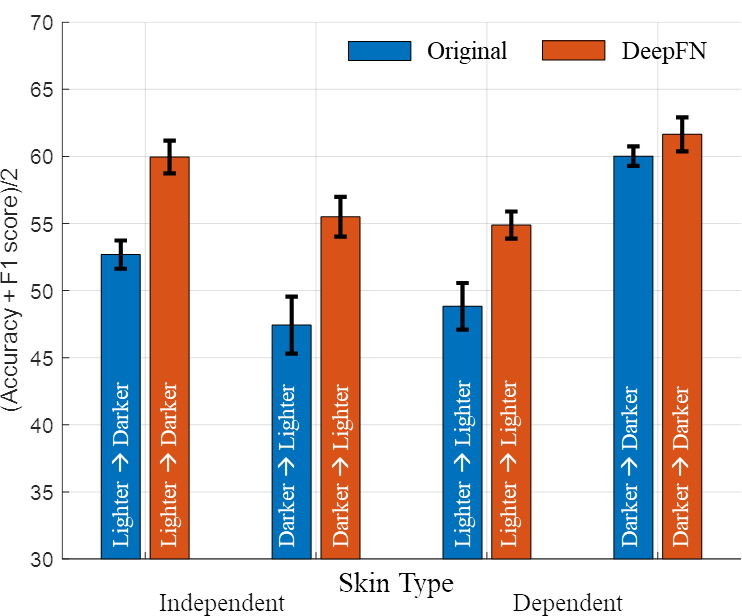}
  \caption{Average and standard error bars for different training/testing splits of skin-independent and dependent models.}
  \label{fig:skin_splits}
\end{figure}

Figure~\ref{fig:skin_splits} shows the results for each of the training/testing combinations with and without DeepFN. We see that the use of DeepFN yielded a consistent average improvement of 5.7\% across the different conditions. This difference was the smallest when training and testing with people with darker skin, in which performance was already at the level of person-dependent models (around 60\%). Similarly, we observe that experiments in which people with darker skin were used as part of the testing set yielded better performance than those considering lighter skin, irrespective of the training group and normalization method (\mbox{$p <= 0.020$}). This finding is consistent with the gender differences as the majority of participants with darker skin were female (6 out of 8) and can be also observed on the average facial appearance of Figure~\ref{fig:avg_face}.

\subsection{Cross-dataset Generalization}

When using the original data, dataset-independent models achieved an average score of 44.6\% and dataset-dependent models achieved an average score of 54\% which were significantly different (\mbox{$p < 0.001$}). This difference indicates that the impact of having different datasets across training and testing sets is around 9.4\%. This difference was the largest observed gap across all the group conditions indicating that cross-dataset generalization is one of the most difficult challenges to address.

When using DeepFN, we observe that dataset-independent models increased to 51.1\% which was significantly higher than the unnormalized counterpart (\mbox{$p < 0.001$}) and dataset-dependent models increased to 61.5\% which was also significantly higher than the unnormalized counterpart (\mbox{$p < 0.001$}). However, the gap between dataset-independent models and dataset-dependent models was significant (\mbox{$p < 0.001$}) indicating that even though DeepFN was able to close a significant part of the gap, there are potentially other sources of unaddressed variance such as illumination and camera angle as shown in Figure~\ref{fig:avg_face}~(right) that could help further normalize facial appearance. 

\begin{figure}[t]
  \centering
  \includegraphics[width=1\linewidth]{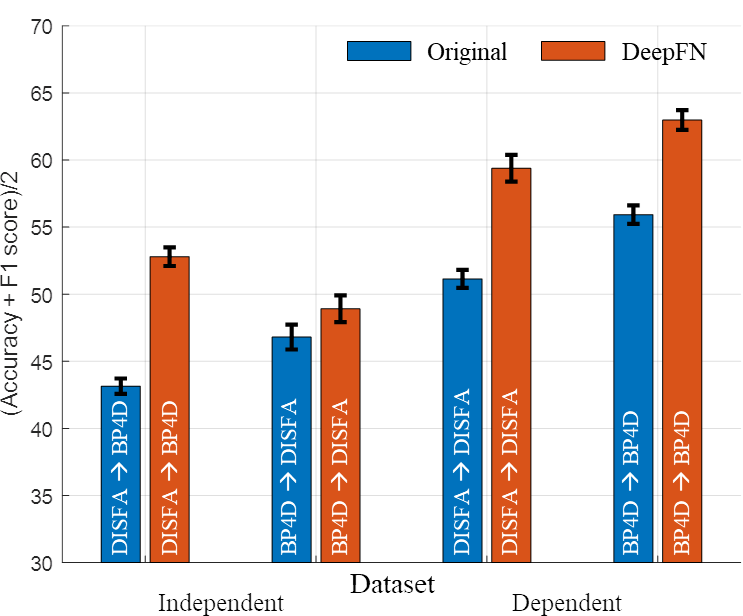}
  \caption{Average and standard error bars for different training/testing splits of dataset-independent and dependent models.}
  \label{fig:db_splits}
\end{figure}

Figure~\ref{fig:db_splits} shows the results for each of the training/testing combinations with and without DeepFN. Note that the right-most condition (i.e.,~BP4D $\rightarrow$ BP4D) is equivalent to the person-independent models considered in section~\ref{subsec:person} but considering a smaller set of facial action units, leading to a slightly higher score. In this condition, within-group evaluations yielded significantly better performance than cross-group evaluations as hypothesized (\mbox{$p < 0.012$}). The use of DeepFN also yielded average improvements of 8.3\% for three of the conditions but only 2.1\% when training on BP4D and testing on DISFA. We believe the smaller improvement may be due to two main factors. On the one hand, the difference in class priors across the two datasets may have limited the ability to learn facial action units. On the other hand, the template of reference was selected from the same distribution as the training data (BP4D) and negatively impacted the expression transfer of more dissimilar images (DISFA). The latter seems to be supported by Figure~\ref{fig:avg_face} that shows greater differences between the original (middle) and normalized (bottom) DISFA images.

\section{Discussion}

This work has proposed and evaluated DeepFN, a deep facial appearance normalization method that minimizes appearance differences across people to help facilitate the task of facial action unit recognition. In particular, we explored the use of self-supervised denoising autoencoders that enable us to transfer facial expressions in a self-supervised manner. 
To evaluate the potential impact of DeepFN in the context of generalization, we performed multiple within-group and cross-group evaluations with both unnormalized and normalized data.

When considering the optimal machine learning scenario in which both labels and images of the testing subjects were available (i.e.,~person-dependent models), we obtained an average F1 and accuracy performance of 60.3\%. In contrast, person-independent models with unnormalized data yielded a significantly lower performance of 55\%, which was successfully corrected when normalizing the data with DeepFN (59.6\%). To analyze different types of person-independent models, we further constrained the groups of participants selected for training and testing sets in terms of gender (male and female), skin type~(lighter and darker), and dataset (BP4D and DISFA). Overall, we observed that cross-group evaluations performed worse than within-group evaluations on average which is consistent with the iid assumption that requires that both training and testing data follow the same distribution. The gap between within and cross-evaluations was the largest for the evaluations examining different datasets (9.4\%) followed by skin type~(5.3\%) and gender (2.4\%). For all the cases, DeepFN helped close a significant part of the gap which was more pronounced when considering gender and skin type.

The previous results seem to support that the use of DeepFN can effectively minimize individual differences while keeping relevant facial expression information. However, we observed some differences across certain conditions that suggest that further improvements could be made in terms of template selection. In our study, we selected the most expressive subject of the BP4D dataset as a template, which helped capture a large range of facial action units. However, the specific attributes of the selected subject more closely resemble certain groups of subjects which could have differently impacted others (e.g.,~Asian female). For instance, we observed that testing on female subjects and people with darker skin, which were predominantly female, led to slightly better performance than its male and lighter skin counterparts. While the use of DeepFN helped slightly close this gap, the differences across groups still existed. In addition, we observed that transferring faces of the DISFA dataset, which was collected under different settings, led to an observable decrease in transfer quality (see~right images of Figure~\ref{fig:avg_face}) which could be expected due to the additional sources of variance (e.g.,~camera angle, illumination) which were not present in the template of reference. 
Future efforts will need to systematically study the role of template selection to enable more consistent normalization results across different groups. We believe the appearance of the ideal template should have similar resemblance across all the groups (e.g.,~average face) as well as capture a rich range of different sources of variance (e.g.,~facial expressions, illumination, body poses). Recent computer vision efforts such as controllable avatars or generative adversarial networks (e.g.,~\cite{Aneja2018,Aneja2019,Tewari2020,VanderStruijk2018}) could be helpful in this space. In addition, we believe there exist some complementary efforts in the context of face frontalization (e.g.,~\cite{Qian2019}) that could be used in combination with the proposed approach to help provide a more comprehensive facial appearance normalization. 


To facilitate facial expression transfer in a self-supervised manner, this work iteratively transferred the data from each person to the template of reference. This approach yielded good qualitative performance but the training process can be considered too temporally costly for real-time analysis. While not explored in this work, there exist multiple methods that can be used to help speed-up the learning process such a re-using an existing pre-trained autoencoder and fine-tuning it with the new data, and/or building a supervised version of the autoencoder that allows normalizing any face without the need of seeing unlabeled data. Training the supervised approach would require obtaining large amounts of paired images (original and normalized faces) which could be provided with the method discussed in this work and/or some of the recent efforts focused on systematic manipulation of synthetic faces (e.g.,~\cite{Nie2020,Tewari2020,Deng2020}). Due to the quick advancements of computer vision, we believe the temporal cost associated with training DeepFN will be significantly reduced in the near term.

To help evaluate the performance of DeepFN across different conditions, this work has made some experimental decisions that are important to consider. For instance, this work has mainly considered evaluations using the BP4D dataset which provided a rich collection of images from different subjects required for the normalization. In addition, the dataset contains a very diverse set of participants in terms of demographics which facilitated exploring different generalization across different groups. 
However, the number of available subjects for some of the groups was relatively small (7 for the people with lighter skin) which constrained the training size across all the other conditions for consistency purposes. This work has also considered the LeNet-5 network to recognize facial action units which facilitated performing a large number of experiments and repetitions (see~Reps. on Table~\ref{tab:overview_results}) in a manageable amount of time, However, it is important to note that this network is relatively simple and that there is a wide variety of more complex network architectures which are currently being used to obtain state-of-the-art-performance (e.g.,~\cite{niu2019local,li2019semantic}). Due to the exploratory nature of this work, we believe these experimental decisions were necessary to facilitate a fair and comprehensive evaluation but acknowledge that  they could have impacted the results. For instance, we would expect to see greater DeepFN benefits when reducing the number of training subjects as well as reducing the complexity of the network, as it may be more difficult to learn generalizable features. In contrast, we may expect decreased benefits when significantly increasing the number of training subjects as well as increasing the complexity of the neural network. While future efforts will need to consider different experimental decisions to further understand the role of DeepFN, especially when considering state-of-the-art methods, we believe the results of this work highlight the potential value of the proposed methodology.

\section{Ethical Considerations}

This work studies the problem of model generalization across different groups which is important to not only improve performance when deploying the models in the real-world but also to help prevent potential biases that can differently impact underrepresented groups (e.g.,~security screening, job interviews). While this work has shown significant performance improvement when considering different groups, it is important to note that there were still some significant gaps across different data splits (e.g.,~testing with female subjects yielded better performance) which should be further minimized to help prevent algorithmic biases. In addition, the evaluations presented in this work were performed in semi-controlled datasets which suffer from their own biases (e.g.,~young population) and may not be representative of real-world demographics. To address these limitations, it is important to consider not only recent advancements in computer vision such as the generation of synthetic images~\cite{Tewari2020} but also acknowledge their unique potential biases~\cite{Salminen2020} which can indirectly increase some of the gaps.

Another challenge when deploying AI systems like the one considered in this work, is the ability to interpret and understand what the models are doing. In contrast to prior work that frequently considered normalizing features with intuitive methods (e.g.,~range correction, relative changes), this work separates the learning process into two main phases: one fully dedicated towards minimizing individual differences, and another one fully dedicated towards learning facial action unit recognition. This separation offers an opportunity to examine the output of the models after the normalization process which can be used to isolate potential failures in the generalization process. Furthermore, the use of a shared facial appearance provided a familiar channel of model introspection (faces) that facilitates the intuitive detection of AI limitations. For instance, Figure~\ref{fig:avg_face} easily allowed us to identify potential limitations when transferring faces across datasets~(right). In the future, we expect to see more approaches further separating the learning process with the hope of better isolating and debugging potential failures.

Finally, it is important to note that facial expression transfer algorithms like the one considered in this work can pose serious risks to the privacy of individuals as their appearance could be easily manipulated without their consent. For instance, there has been an increase of manipulated videos in sensitive scenarios such as politics and pornography with very profound and worrisome implications to society~\cite{Mirsky2020}. While there has been recent progress towards the automated detection of these misuses (e.g.,~\cite{Rossler2019,Guera2019,Ciftci2020}), it is critical to find ways to obtain the informed consent of users. In this work, we mitigated this by only considering the appearance of participants who explicitly contributed their data for research purposes as a template of reference. However, future work considering different template selections will need to consider the importance of obtaining consent too. 
Despite the shortcomings of expression transfer methods, however, we believe there are several positive applications such as helping preserve the privacy of individuals (e.g.,~\cite{Li2019,Hukkelas2019,Zhu2020}) and helping minimize potential algorithmic biases as explored in this work.


\section{Conclusion}
This work has explored the use of facial expression transfer to minimize individual differences across people in the context of facial action unit recognition, and systematically evaluated its potential generalization benefits by performing multiple within and cross-group comparisons in terms of people, gender, skin type, and dataset. The results of this work demonstrate that the proposed methodology can yield significant generalization gains but more work is needed to assess its replicability across different experimental conditions. We are looking forward to a future when similar face normalization methods can be used not only to deploy more accurate models in the wild but also provide more consistent performance across different groups of people.


%





\ifCLASSOPTIONcaptionsoff
  \newpage
\fi



%




\bibliographystyle{unsrt}
\bibliography{sample-base}

\begin{thebibliography}{10}
\providecommand{\url}[1]{#1}
\csname url@samestyle\endcsname
\providecommand{\newblock}{\relax}
\providecommand{\bibinfo}[2]{#2}
\providecommand{\BIBentrySTDinterwordspacing}{\spaceskip=0pt\relax}
\providecommand{\BIBentryALTinterwordstretchfactor}{4}
\providecommand{\BIBentryALTinterwordspacing}{\spaceskip=\fontdimen2\font plus
\BIBentryALTinterwordstretchfactor\fontdimen3\font minus
  \fontdimen4\font\relax}
\providecommand{\BIBforeignlanguage}[2]{{%
\expandafter\ifx\csname l@#1\endcsname\relax
\typeout{** WARNING: IEEEtran.bst: No hyphenation pattern has been}%
\typeout{** loaded for the language `#1'. Using the pattern for}%
\typeout{** the default language instead.}%
\else
\language=\csname l@#1\endcsname
\fi
#2}}
\providecommand{\BIBdecl}{\relax}
\BIBdecl

\bibitem{Martinez2019}
B.~Martinez, M.~F. Valstar, B.~Jiang, and M.~Pantic, ``{Automatic analysis of
  facial actions: A survey},'' pp. 325--347, jul 2019.

\bibitem{barrett2019emotional}
L.~F. Barrett, R.~Adolphs, S.~Marsella, A.~M. Martinez, and S.~D. Pollak,
  ``Emotional expressions reconsidered: Challenges to inferring emotion from
  human facial movements,'' \emph{Psychological science in the public
  interest}, vol.~20, no.~1, pp. 1--68, 2019.

\bibitem{Rudovic2018a}
O.~Rudovic, Y.~Utsumi, J.~Lee, J.~Hernandez, E.~C. Ferrer, B.~Schuller, and
  R.~W. Picard, ``{CultureNet: A Deep Learning Approach for Engagement
  Intensity Estimation from Face Images of Children with Autism},'' \emph{IEEE
  International Conference on Intelligent Robots and Systems}, no. August, pp.
  339--346, 2018.

\bibitem{Rudovic2018b}
O.~Rudovic, J.~Lee, M.~Dai, B.~Schuller, and R.~W. Picard, ``{Personalized
  machine learning for robot perception of affect and engagement in autism
  therapy},'' \emph{Science Robotics}, vol.~3, no.~19, 2018.

\bibitem{Cohn2009}
J.~F. Cohn, T.~S. Kruez, I.~Matthews, Y.~Yang, M.~H. Nguyen, M.~T. Padilla,
  F.~Zhou, and F.~{De La Torre}, ``{Detecting depression from facial actions
  and vocal prosody},'' in \emph{Proceedings - 2009 3rd International
  Conference on Affective Computing and Intelligent Interaction and Workshops,
  ACII 2009}, 2009.

\bibitem{stratou2013automatic}
G.~Stratou, S.~Scherer, J.~Gratch, and L.-P. Morency, ``Automatic nonverbal
  behavior indicators of depression and ptsd: Exploring gender differences,''
  in \emph{2013 Humaine Association Conference on Affective Computing and
  Intelligent Interaction}.\hskip 1em plus 0.5em minus 0.4em\relax IEEE, 2013,
  pp. 147--152.

\bibitem{Lucey2011}
P.~Lucey, J.~F. Cohn, I.~Matthews, S.~Lucey, S.~Sridharan, J.~Howlett, and
  K.~M. Prkachin, ``{Automatically detecting pain in video through facial
  action units.}'' \emph{IEEE transactions on systems, man, and cybernetics.
  Part B, Cybernetics : a publication of the IEEE Systems, Man, and Cybernetics
  Society}, vol.~41, no.~3, pp. 664--74, jun 2011.

\bibitem{kaltwang2012continuous}
S.~Kaltwang, O.~Rudovic, and M.~Pantic, ``Continuous pain intensity estimation
  from facial expressions,'' in \emph{International Symposium on Visual
  Computing}.\hskip 1em plus 0.5em minus 0.4em\relax Springer, 2012, pp.
  368--377.

\bibitem{martinez2017automatic}
B.~Martinez, M.~F. Valstar, B.~Jiang, and M.~Pantic, ``Automatic analysis of
  facial actions: A survey,'' \emph{IEEE transactions on affective computing},
  2017.

\bibitem{McDuff2015}
D.~McDuff, R.~E. Kaliouby, J.~F. Cohn, and R.~W. Picard, ``{Predicting Ad
  Liking and Purchase Intent: Large-Scale Analysis of Facial Responses to
  Ads},'' \emph{IEEE Transactions on Affective Computing}, vol.~6, no.~3, pp.
  223--235, jul 2015.

\bibitem{Hernandez2013b}
J.~Hernandez, Z.~Liu, G.~Hulten, D.~Debarr, K.~Krum, and Z.~Zhang, ``{Measuring
  the engagement level of TV viewers},'' in \emph{2013 10th IEEE International
  Conference and Workshops on Automatic Face and Gesture Recognition, FG 2013},
  2013.

\bibitem{Assari2011}
M.~A. Assari and M.~Rahmati, ``{Driver drowsiness detection using face
  expression recognition},'' in \emph{2011 IEEE International Conference on
  Signal and Image Processing Applications, ICSIPA 2011}, 2011, pp. 337--341.

\bibitem{Gao2014}
H.~Gao, A.~Yuce, and J.~P. Thiran, ``{Detecting emotional stress from facial
  expressions for driving safety},'' in \emph{2014 IEEE International
  Conference on Image Processing, ICIP 2014}.\hskip 1em plus 0.5em minus
  0.4em\relax Institute of Electrical and Electronics Engineers Inc., jan 2014,
  pp. 5961--5965.

\bibitem{ekman1997face}
R.~Ekman, \emph{What the face reveals: Basic and applied studies of spontaneous
  expression using the Facial Action Coding System (FACS)}.\hskip 1em plus
  0.5em minus 0.4em\relax Oxford University Press, USA, 1997.

\bibitem{Cohn2007}
J.~F. Cohn, Z.~Ambadar, and P.~Ekman, ``{Observer-based measurement of facial
  expression with the Facial Action Coding System},'' in \emph{Series in
  affective science. Handbook of emotion elicitation and assessment}, {J. A.
  Coan {\&} J. J. B. Allen}, Ed.\hskip 1em plus 0.5em minus 0.4em\relax Oxford
  University Press, 2007, ch. Observer-b, pp. 203--221.

\bibitem{zhao2018learning}
K.~Zhao, W.-S. Chu, and A.~M. Martinez, ``Learning facial action units from web
  images with scalable weakly supervised clustering,'' in \emph{Proceedings of
  the IEEE Conference on Computer Vision and Pattern Recognition}, 2018, pp.
  2090--2099.

\bibitem{li2019semantic}
G.~Li, X.~Zhu, Y.~Zeng, Q.~Wang, and L.~Lin, ``Semantic relationships guided
  representation learning for facial action unit recognition,'' in
  \emph{Proceedings of the AAAI Conference on Artificial Intelligence},
  vol.~33, 2019, pp. 8594--8601.

\bibitem{baltrusaitis2018openface}
T.~Baltrusaitis, A.~Zadeh, Y.~C. Lim, and L.-P. Morency, ``Openface 2.0: Facial
  behavior analysis toolkit,'' in \emph{2018 13th IEEE International Conference
  on Automatic Face \& Gesture Recognition (FG 2018)}.\hskip 1em plus 0.5em
  minus 0.4em\relax IEEE, 2018, pp. 59--66.

\bibitem{mcduff2019multimodal}
D.~McDuff, K.~Rowan, P.~Choudhury, J.~Wolk, T.~Pham, and M.~Czerwinski, ``A
  multimodal emotion sensing platform for building emotion-aware
  applications,'' \emph{arXiv preprint arXiv:1903.12133}, 2019.

\bibitem{Corneanu2016}
C.~A. Corneanu, M.~O. Sim{\'{o}}n, J.~F. Cohn, and S.~E. Guerrero, ``{Survey on
  RGB, 3D, Thermal, and Multimodal Approaches for Facial Expression
  Recognition: History, Trends, and Affect-Related Applications},'' \emph{IEEE
  Transactions on Pattern Analysis and Machine Intelligence}, vol.~38, no.~8,
  pp. 1548--1568, aug 2016.

\bibitem{Li2020}
S.~Li and W.~Deng, ``{Deep Facial Expression Recognition: A Survey},''
  \emph{IEEE Transactions on Affective Computing}, 2020.

\bibitem{Cohen2003}
I.~Cohen, N.~Sebe, A.~Garg, L.~S. Chen, and T.~S. Huang, ``{Facial expression
  recognition from video sequences: Temporal and static modeling},''
  \emph{Computer Vision and Image Understanding}, vol.~91, no. 1-2, pp.
  160--187, jul 2003.

\bibitem{Valstar2011}
M.~F. Valstar, B.~Jiang, M.~Mehu, M.~Pantic, and K.~Scherer, ``{The first
  facial expression recognition and analysis challenge},'' in \emph{2011 IEEE
  International Conference on Automatic Face and Gesture Recognition and
  Workshops, FG 2011}, 2011, pp. 921--926.

\bibitem{Fink2005}
B.~Fink, K.~Grammer, P.~Mitteroecker, P.~Gunz, K.~Schaefer, F.~L. Bookstein,
  and J.~T. Manning, ``{Second to fourth digit ratio and face shape},''
  \emph{Proceedings of the Royal Society B: Biological Sciences}, vol. 272, no.
  1576, pp. 1995--2001, oct 2005.

\bibitem{Zhuang2010}
Z.~Zhuang, D.~Landsittel, S.~Benson, R.~Roberge, and R.~Shaffer, ``{Facial
  anthropometric differences among gender, ethnicity, and age groups},''
  \emph{Annals of Occupational Hygiene}, vol.~54, no.~4, pp. 391--402, jun
  2010.

\bibitem{buolamwini2017gender}
J.~A. Buolamwini, ``Gender shades: intersectional phenotypic and demographic
  evaluation of face datasets and gender classifiers,'' Ph.D. dissertation,
  Massachusetts Institute of Technology, 2017.

\bibitem{buolamwini2018gender}
J.~Buolamwini and T.~Gebru, ``Gender shades: Intersectional accuracy
  disparities in commercial gender classification,'' in \emph{Conference on
  Fairness, Accountability and Transparency}, 2018, pp. 77--91.

\bibitem{wilson2019predictive}
B.~{Wilson}, J.~{Hoffman}, and J.~{Morgenstern}, ``{Predictive Inequity in
  Object Detection},'' \emph{arXiv e-prints}, p. arXiv:1902.11097, Feb 2019.

\bibitem{Quattrone1980}
G.~A. Quattrone and E.~E. Jones, ``{The perception of variability within
  in-groups and out-groups: Implications for the law of small numbers.}''
  \emph{Journal of Personality and Social Psychology}, vol.~38, no.~1, pp.
  141--152, 1980.

\bibitem{Elfenbein2002}
H.~A. Elfenbein and N.~Ambady, ``{On the universality and cultural specificity
  of emotion recognition: A meta-analysis},'' pp. 203--235, 2002.

\bibitem{Chen2013}
J.~Chen, X.~Liu, P.~Tu, and A.~Aragones, ``{Learning person-specific models for
  facial expression and action unit recognition},'' \emph{Pattern Recognition
  Letters}, vol.~34, no.~15, pp. 1964--1970, 2013.

\bibitem{Tian2001}
Y.~L. Tian, T.~Kanade, and J.~F. Conn, ``{Recognizing action units for facial
  expression analysis},'' \emph{IEEE Transactions on Pattern Analysis and
  Machine Intelligence}, vol.~23, no.~2, pp. 97--115, 2001.

\bibitem{hernandez2013measuring}
J.~Hernandez, Z.~Liu, G.~Hulten, D.~DeBarr, K.~Krum, and Z.~Zhang, ``Measuring
  the engagement level of tv viewers,'' in \emph{2013 10th IEEE International
  Conference and Workshops on Automatic Face and Gesture Recognition
  (FG)}.\hskip 1em plus 0.5em minus 0.4em\relax IEEE, 2013, pp. 1--7.

\bibitem{Niu2019}
X.~Niu, H.~Han, S.~Yang, Y.~Huang, and S.~Shan, ``{Local relationship learning
  with person-specific shape regularization for facial action unit
  detection},'' \emph{Proceedings of the IEEE Computer Society Conference on
  Computer Vision and Pattern Recognition}, vol. 2019-June, pp.
  11\,909--11\,918, 2019.

\bibitem{Tran2017}
D.~L. Tran, R.~Walecki, Ognjen, Rudovic, S.~Eleftheriadis, B.~Schuller, and
  M.~Pantic, ``{DeepCoder: Semi-parametric Variational Autoencoders for Facial
  Action Unit Intensity Estimation},'' pp. 3190--3199, 2017.

\bibitem{banerjee2018frontalize}
S.~Banerjee, J.~Brogan, J.~Krizaj, A.~Bharati, B.~R. Webster, V.~Struc, P.~J.
  Flynn, and W.~J. Scheirer, ``To frontalize or not to frontalize: Do we really
  need elaborate pre-processing to improve face recognition?'' in \emph{2018
  IEEE Winter Conference on Applications of Computer Vision (WACV)}.\hskip 1em
  plus 0.5em minus 0.4em\relax IEEE, 2018, pp. 20--29.

\bibitem{Baltrusaitis2015}
T.~Baltru{\v{s}}aitis, M.~Mahmoud, and P.~Robinson, ``{Cross-dataset learning
  and person-specific normalisation for automatic Action Unit detection},''
  \emph{2015 11th IEEE International Conference and Workshops on Automatic Face
  and Gesture Recognition, FG 2015}, vol. 2015-Janua, 2015.

\bibitem{Chu2017}
W.~S. Chu, F.~{De La Torre}, and J.~F. Cohn, ``{Selective transfer machine for
  personalized facial expression analysis},'' \emph{IEEE Transactions on
  Pattern Analysis and Machine Intelligence}, vol.~39, no.~3, pp. 529--545,
  2017.

\bibitem{zen2014unsupervised}
G.~Zen, E.~Sangineto, E.~Ricci, and N.~Sebe, ``Unsupervised domain adaptation
  for personalized facial emotion recognition,'' in \emph{Proceedings of the
  16th international conference on multimodal interaction}, 2014, pp. 128--135.

\bibitem{feffer2018mixture}
M.~Feffer, R.~W. Picard \emph{et~al.}, ``A mixture of personalized experts for
  human affect estimation,'' in \emph{International Conference on Machine
  Learning and Data Mining in Pattern Recognition}.\hskip 1em plus 0.5em minus
  0.4em\relax Springer, 2018, pp. 316--330.

\bibitem{yang2014personalized}
S.~Yang, O.~Rudovic, V.~Pavlovic, and M.~Pantic, ``Personalized modeling of
  facial action unit intensity,'' in \emph{International Symposium on Visual
  Computing}.\hskip 1em plus 0.5em minus 0.4em\relax Springer, 2014, pp.
  269--281.

\bibitem{qian2019unsupervised}
Y.~Qian, W.~Deng, and J.~Hu, ``Unsupervised face normalization with extreme
  pose and expression in the wild,'' in \emph{Proceedings of the IEEE
  Conference on Computer Vision and Pattern Recognition}, 2019, pp. 9851--9858.

\bibitem{haghighat2016fully}
M.~Haghighat, M.~Abdel-Mottaleb, and W.~Alhalabi, ``Fully automatic face
  normalization and single sample face recognition in unconstrained
  environments,'' \emph{Expert Systems with Applications}, vol.~47, pp. 23--34,
  2016.

\bibitem{sagonas2017robust}
C.~Sagonas, Y.~Panagakis, S.~Zafeiriou, and M.~Pantic, ``Robust statistical
  frontalization of human and animal faces,'' \emph{International journal of
  computer vision}, vol. 122, no.~2, pp. 270--291, 2017.

\bibitem{Werner2019}
P.~Werner, F.~Saxen, A.~Al-Hamadi, and H.~Yu, ``{Generalizing to unseen head
  poses in facial expression recognition and action unit intensity
  estimation},'' \emph{Proceedings - 14th IEEE International Conference on
  Automatic Face and Gesture Recognition, FG 2019}, 2019.

\bibitem{Thies2015}
J.~Thies, M.~Zollh{\"{o}}fer, M.~Nie{\ss}ner, L.~Valgaerts, M.~Stamminger, and
  C.~Theobalt, ``{Real-time expression transfer for facial reenactment},''
  \emph{ACM Transactions on Graphics}, vol.~34, no.~6, 2015.

\bibitem{Kim2018}
H.~Kim, P.~Garrido, A.~Tewari, W.~Xu, J.~Thies, M.~Niessner, P.~P{\'{e}}rez,
  C.~Richardt, M.~Zollh{\"{o}}fer, and C.~Theobalt, ``{Deep video portraits},''
  \emph{ACM Transactions on Graphics}, vol.~37, no.~4, 2018.

\bibitem{Zakharov2019}
E.~Zakharov, A.~Shysheya, E.~Burkov, and V.~Lempitsky, ``{Few-shot adversarial
  learning of realistic neural talking head models},'' \emph{Proceedings of the
  IEEE International Conference on Computer Vision}, vol. 2019-Octob, no. May,
  pp. 9458--9467, 2019.

\bibitem{Nirkin2019}
Y.~Nirkin, Y.~Keller, and T.~Hassner, ``{FSGAN: Subject agnostic face swapping
  and reenactment},'' \emph{Proceedings of the IEEE International Conference on
  Computer Vision}, vol. 2019-Octob, pp. 7183--7192, 2019.

\bibitem{Mirsky2020}
Y.~Mirsky and W.~Lee, ``{The Creation and Detection of Deepfakes: A Survey},''
  \emph{ACM Comput. Surv. 1, 1, Article}, vol.~1, apr 2020.

\bibitem{VanderStruijk2018}
S.~van~der Struijk, M.~S. Mirzaei, H.~H. Huang, and T.~Nishida, ``{Facsvatar:
  An Open Source Modular Framework for Real-Time FACS based Facial
  Animation},'' in \emph{Proceedings of the 18th International Conference on
  Intelligent Virtual Agents, IVA 2018}.\hskip 1em plus 0.5em minus 0.4em\relax
  New York, NY, USA: Association for Computing Machinery, Inc, nov 2018, pp.
  159--164.

\bibitem{Aneja2019}
D.~Aneja, D.~McDuff, and S.~Shah, ``{A high-fidelity open embodied avatar with
  lip syncing and expression capabilities},'' in \emph{ICMI 2019 - Proceedings
  of the 2019 International Conference on Multimodal Interaction}.\hskip 1em
  plus 0.5em minus 0.4em\relax New York, NY, USA: Association for Computing
  Machinery, Inc, oct 2019, pp. 69--73.

\bibitem{Karras2019}
T.~Karras, S.~Laine, and T.~Aila, ``{A style-based generator architecture for
  generative adversarial networks},'' in \emph{Proceedings of the IEEE Computer
  Society Conference on Computer Vision and Pattern Recognition}, vol.
  2019-June.\hskip 1em plus 0.5em minus 0.4em\relax IEEE Computer Society, jun
  2019, pp. 4396--4405.

\bibitem{Salminen2020}
J.~Salminen, S.~G. Jung, S.~Chowdhury, and B.~J. Jansen, ``{Analyzing
  demographic bias in artificially generated facial pictures},''
  \emph{Conference on Human Factors in Computing Systems - Proceedings}, pp.
  1--8, 2020.

\bibitem{LeCun1998}
Y.~LeCun, L.~Bottou, Y.~Bengio, and P.~Haffner, ``{Gradient-based learning
  applied to document recognition},'' \emph{Proceedings of the IEEE}, vol.~86,
  no.~11, pp. 2278--2323, 1998.

\bibitem{Qian2019}
Y.~Qian, W.~Deng, and J.~Hu, ``{Unsupervised face normalization with extreme
  pose and expression in the wild},'' \emph{Proceedings of the IEEE Computer
  Society Conference on Computer Vision and Pattern Recognition}, vol.
  2019-June, pp. 9843--9850, 2019.

\bibitem{fitzpatrick1988validity}
T.~B. Fitzpatrick, ``The validity and practicality of sun-reactive skin types i
  through vi,'' \emph{Archives of dermatology}, vol. 124, no.~6, pp. 869--871,
  1988.

\bibitem{Zhang2014}
X.~Zhang, L.~Yin, J.~F. Cohn, S.~Canavan, M.~Reale, A.~Horowitz, P.~Liu, and
  J.~M. Girard, ``{BP4D-Spontaneous: A high-resolution spontaneous 3D dynamic
  facial expression database},'' \emph{Image and Vision Computing}, vol.~32,
  no.~10, pp. 692--706, oct 2014.

\bibitem{Mavadati2013}
S.~M. Mavadati, M.~H. Mahoor, K.~Bartlett, P.~Trinh, and J.~F. Cohn, ``{DISFA:
  A spontaneous facial action intensity database},'' \emph{IEEE Transactions on
  Affective Computing}, vol.~4, no.~2, pp. 151--160, 2013.

\bibitem{Kanade2000}
T.~Kanade, J.~F. Cohn, and Y.~Tian, ``{Comprehensive database for facial
  expression analysis},'' in \emph{Proceedings - 4th IEEE International
  Conference on Automatic Face and Gesture Recognition, FG 2000}, 2000.

\bibitem{Aneja2018}
D.~Aneja, B.~Chaudhuri, A.~Colburn, G.~Faigin, L.~Shapiro, and B.~Mones,
  ``{Learning to Generate 3D Stylized Character Expressions from Humans},''
  \emph{Proceedings - 2018 IEEE Winter Conference on Applications of Computer
  Vision, WACV 2018}, vol. 2018-Janua, pp. 160--169, 2018.

\bibitem{Tewari2020}
A.~Tewari, M.~Elgharib, G.~Bharaj, F.~Bernard, H.-P. Seidel, P.~Perez,
  M.~Zollhofer, and C.~Theobalt, ``{StyleRig: Rigging StyleGAN for 3D Control
  Over Portrait Images},'' pp. 6141--6150, 2020.

\bibitem{Nie2020}
W.~Nie, T.~Karras, A.~Garg, S.~Debnath, A.~Patney, A.~B. Patel, and
  A.~Anandkumar, ``{Semi-Supervised StyleGAN for Disentanglement Learning},''
  in \emph{International Conference on Machine Learning (ICML)}, mar 2020.

\bibitem{Deng2020}
Y.~Deng, J.~Yang, D.~Chen, F.~Wen, and X.~Tong, ``{Disentangled and
  Controllable Face Image Generation via 3D Imitative-Contrastive Learning
  Generated image Pose Expression Illumination Random corpus},'' in
  \emph{Proceedings of the IEEE/CVF Conference on Computer Vision and Pattern
  Recognition (CVPR)}, jun 2020, pp. 5154--5163.

\bibitem{niu2019local}
X.~Niu, H.~Han, S.~Yang, Y.~Huang, and S.~Shan, ``Local relationship learning
  with person-specific shape regularization for facial action unit detection,''
  in \emph{Proceedings of the IEEE Conference on Computer Vision and Pattern
  Recognition}, 2019, pp. 11\,917--11\,926.

\bibitem{Rossler2019}
A.~Rossler, D.~Cozzolino, L.~Verdoliva, C.~Riess, J.~Thies, and M.~Niessner,
  ``{FaceForensics++: Learning to detect manipulated facial images},''
  \emph{Proceedings of the IEEE International Conference on Computer Vision},
  vol. 2019-Octob, pp. 1--11, 2019.

\bibitem{Guera2019}
D.~Guera and E.~J. Delp, ``{Deepfake Video Detection Using Recurrent Neural
  Networks},'' \emph{Proceedings of AVSS 2018 - 2018 15th IEEE International
  Conference on Advanced Video and Signal-Based Surveillance}, 2019.

\bibitem{Ciftci2020}
U.~A. Ciftci, I.~Demir, and L.~Yin, ``{FakeCatcher: Detection of Synthetic
  Portrait Videos using Biological Signals},'' \emph{IEEE Transactions on
  Pattern Analysis and Machine Intelligence}, pp. 1--1, 2020.

\bibitem{Li2019}
Y.~Li and S.~Lyu, ``{De-identification without losing faces},'' \emph{IH and
  MMSec 2019 - Proceedings of the ACM Workshop on Information Hiding and
  Multimedia Security}, pp. 83--88, 2019.

\bibitem{Hukkelas2019}
H.~Hukkel{\aa}s, R.~Mester, and F.~Lindseth, ``{DeepPrivacy: A Generative
  Adversarial Network for Face Anonymization},'' \emph{Lecture Notes in
  Computer Science (including subseries Lecture Notes in Artificial
  Intelligence and Lecture Notes in Bioinformatics)}, vol. 11844 LNCS, pp.
  565--578, sep 2019.

\bibitem{Zhu2020}
B.~Zhu, H.~Fang, Y.~Sui, and L.~Li, ``{Deepfakes for medical video
  de-identification: Privacy protection and diagnostic information
  preservation},'' in \emph{AIES 2020 - Proceedings of the AAAI/ACM Conference
  on AI, Ethics, and Society}, vol.~7, no.~20.\hskip 1em plus 0.5em minus
  0.4em\relax New York, NY, USA: Association for Computing Machinery, Inc, feb
  2020, pp. 414--420.

\end{thebibliography}

\end{document}